%% file: main.tex
\documentclass[runningheads]{llncs}

\usepackage[mobile]{eccv}
\usepackage{eccvabbrv}

\usepackage{graphicx}
\usepackage{booktabs}
\usepackage{amsmath,amssymb,amsfonts}
\usepackage{algorithm}
\usepackage{algorithmic}
\usepackage{xcolor}
\usepackage{colortbl}
\definecolor{HeatBest}{RGB}{255,199,206}
\definecolor{HeatSecond}{RGB}{255,229,153}
\newcommand{\best}[1]{\cellcolor{HeatBest}{#1}}
\newcommand{\second}[1]{\cellcolor{HeatSecond}{#1}}

\usepackage[accsupp]{axessibility}

\usepackage{hyperref}
\usepackage{orcidlink}

\begin{document}

\title{DiffProxy: Multi-View Human Mesh Recovery via Diffusion-Generated Dense Proxies}

\titlerunning{DiffProxy: Multi-View Human Mesh Recovery}

\author{
Renke Wang\inst{1} \and
Zhenyu Zhang\textsuperscript{*}\inst{2} \and
Ying Tai\inst{2} \and
Jun Li\textsuperscript{*}\inst{1} \and
Jian Yang\textsuperscript{*}\inst{1}
}

\authorrunning{R.~Wang et al.}

\institute{
PCA Lab\textsuperscript{\dag}, Nanjing University of Science and Technology, Nanjing, China
\and
School of Intelligent Science and Technology, Nanjing University, Nanjing, China
}

\maketitle
\makeatletter
\begingroup
\def\@fnsymbol#1{\ensuremath{\ifcase#1\or\star\or\dagger\else\@ctrerr\fi}}
\renewcommand*{\thefootnote}{\@fnsymbol\c@footnote}
\footnotetext[1]{Corresponding authors.}
\footnotetext[2]{PCA Lab, Key Lab of Intelligent Perception and Systems for High-Dimensional Information of Ministry of Education, and Jiangsu Key Lab of Image and Video Understanding for Social Security, School of Computer Science and Engineering, Nanjing University of Sci.\ \& Tech.}
\endgroup
\makeatother

\noindent
\begin{center}
    \includegraphics[width=\linewidth]{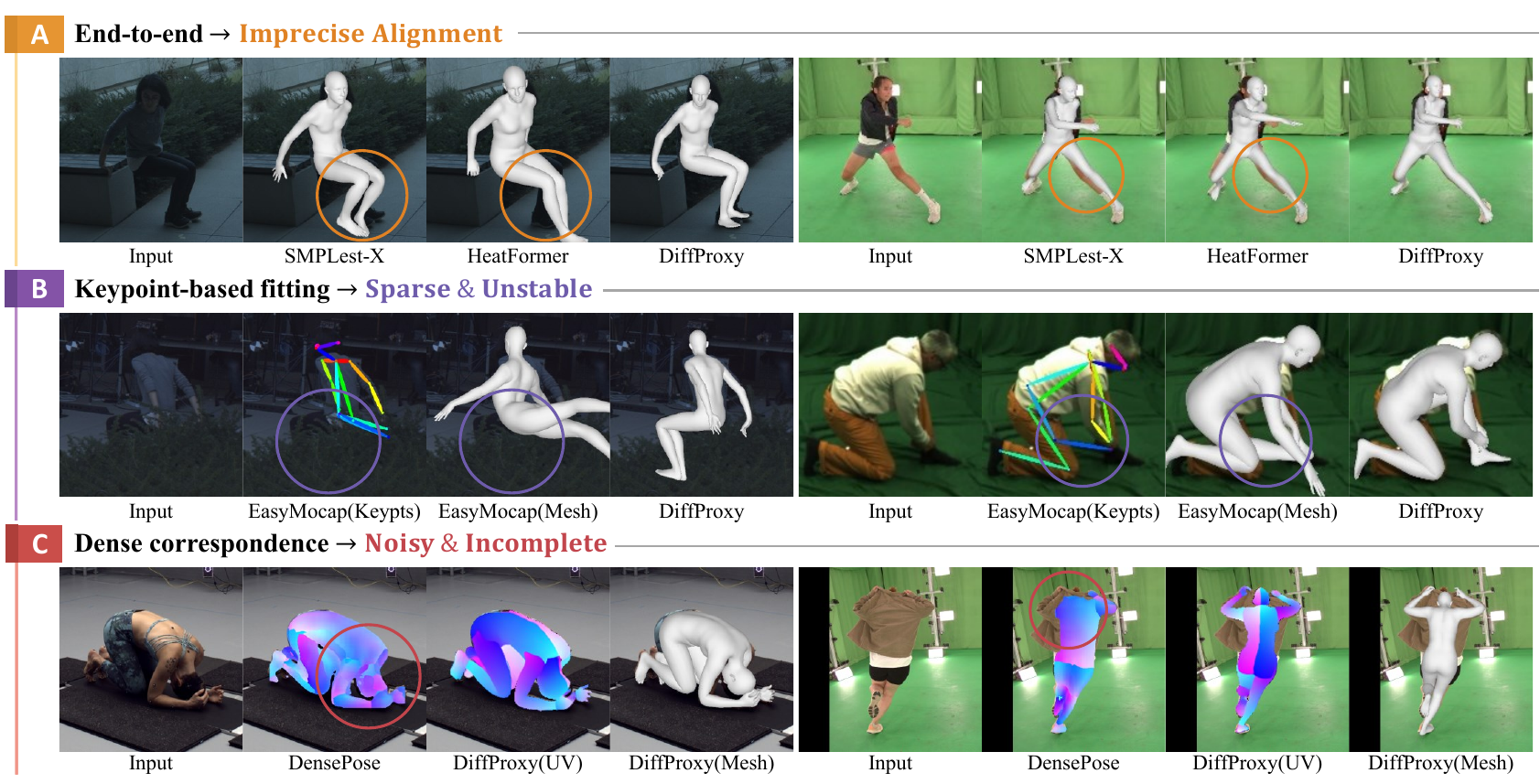}
    \captionsetup{hypcap=false}
    \captionof{figure}{Existing human mesh recovery methods face distinct challenges. \textbf{(A)} End-to-end methods (SMPLest-X, HeatFormer) produce imprecise image-mesh alignment; \textbf{(B)} keypoint-based fitting (EasyMoCap) is sparse and unstable under challenging conditions; \textbf{(C)} prior dense correspondence predictors (DensePose) produce noisy and incomplete surface mappings. We introduce DiffProxy, which uses a diffusion model trained on synthetic data to predict accurate dense pixel-to-surface proxies, then fits SMPL-X against them via a unified reprojection objective, achieving precise alignment across diverse real-world scenarios.}
    \label{fig:teaser}
\end{center}

\input{sec/0_abstract}    
\input{sec/1_introduction}
\input{sec/2_related_work}
\input{sec/3_method}

\input{sec/4_experiments}

\input{sec/5_limitation_and_future_works}
\input{sec/6_conclusion}
\input{sec/7_acknowledgments}

{
    \small
    \bibliographystyle{splncs04}
    \bibliography{main.bib}
}


\end{document}

%% file: sec/0_abstract.tex
\begin{abstract}
    Precise human mesh recovery (HMR) from multi-view images remains challenging: end-to-end methods produce entangled errors hard to localize, while fitting-based methods rely on sparse keypoints that provide limited surface constraints. We observe that the true bottleneck lies in the quality of intermediate representations, and that dense pixel-to-surface correspondences can be effectively generated by repurposing pre-trained diffusion models with rich visual priors. We propose \textbf{DiffProxy}, a Stable-Diffusion-based framework trained on large-scale synthetic data with pixel-perfect annotations. A multi-conditional proxy generator predicts dense correspondences from multi-view images, providing uniform surface constraints that enable precise fitting. Hand refinement feeds enlarged hand crops alongside full-body images for fine-grained detail, while test-time scaling exploits diffusion stochasticity to estimate per-pixel uncertainty. Trained only on synthetic data, DiffProxy achieves state-of-the-art results on five diverse real-world benchmarks. Project page: \url{https://wrk226.github.io/DiffProxy.html}
    \end{abstract}

%% file: sec/1_introduction.tex
\section{Introduction}
\label{sec:introduction}

Human mesh recovery (HMR) aims to reconstruct parametric 3D body models from images and has become a cornerstone of computer vision. The dominant paradigm is \emph{end-to-end prediction}: a neural network directly maps an input image to SMPL/SMPL-X~\cite{SMPL:2015,SMPL-X:2019} parameters in a single forward pass~\cite{lin2021metro,goel2023humans4d,yin2025smplestx,chen2025human3r}. While remarkably fast, this mapping compresses geometry, correspondence, and occlusion reasoning into a single regression target, producing entangled errors that are difficult to localize and often cause image-mesh misalignment~\cite{cheng2023dna,kanazawa2019learning,black2023bedlam}.

Not all applications demand real-time speed. In film production, motion capture annotation~\cite{easymocap}, biomechanical analysis, and digital human creation, precision takes priority. For such scenarios, \emph{fitting-based} methods~\cite{bogo2016smplify,smplifyx,kolotouros2019spin,easymocap} remain the preferred choice, optimizing body model parameters against detected 2D cues. The canonical multi-view pipeline detects 2D keypoints, triangulates them into 3D, and fits the body model against these 3D keypoints. However, sparse keypoints provide limited constraints: a single erroneous detection or occluded joint can dominate the optimization and produce a completely wrong pose (\cref{fig:teaser}).

A natural remedy is to replace sparse keypoints with \emph{dense} pixel-to-surface correspondences, which provide far more constraints per view. Since DensePose~\cite{guler2018densepose} first established dense image-to-surface mappings, several works~\cite{guler2019holopose,xu2019denserac,zeng2020decomr,le2024meshpose} have explored this direction for mesh recovery. Despite the conceptual appeal, these approaches were hindered by two bottlenecks: \textbf{(i)} the DensePose-COCO training set relies on manual annotations ($\sim$50K images) that are inherently noisy, propagating label errors into learned models; \textbf{(ii)} the CNN-based detectors used in these works lack large-scale visual priors and generalize poorly to out-of-distribution scenarios. As a result, predicted correspondences were insufficiently accurate for high-precision fitting, and the dense-proxy route was largely superseded by end-to-end approaches.

We argue that these bottlenecks reflect limitations of the training data and models available at the time, not inherent flaws of the dense-proxy approach. If we can generate sufficiently accurate dense correspondences, the fitting problem simplifies significantly, as dense pixel-to-surface mappings provide far more constraints per view than sparse keypoints, with uniform coverage across the visible body surface. Recent work has shown that pre-trained diffusion models can be repurposed as powerful dense predictors~\cite{ke2024marigold,xu2024matters}, inheriting rich visual priors for cross-domain generalization. This motivates us to revisit the dense-proxy route by addressing both historical bottlenecks: we train on large-scale synthetic data with pixel-perfect SMPL-X correspondences to eliminate label noise, and leverage Stable Diffusion~2.1~\cite{stabilityai2022sd21} for synthetic-to-real transfer.

Building on this foundation, we propose \textbf{DiffProxy}, a framework for precise multi-view human mesh recovery (\cref{fig:teaser}). At its core, a \emph{multi-conditional proxy generator} predicts dense pixel-to-surface correspondences from multi-view images, and SMPL-X parameters are recovered by fitting against them via a differentiable reprojection objective. We further introduce \emph{hand refinement}---since hands occupy too few pixels for reliable full-body prediction, we crop and enlarge hand regions as additional views for fine-grained detail via cross-view attention---and \emph{test-time scaling}---for challenging regions such as occlusions or uncommon poses, individual predictions can be erroneous, so we aggregate multiple stochastic samples to derive per-pixel uncertainty for down-weighting unreliable correspondences during fitting. These form a unified pipeline trained entirely on synthetic data.

In summary, our main contributions are:
\begin{itemize}
    \item We construct a large-scale synthetic multi-view dataset (105K samples, 844K images) with pixel-perfect SMPL-X annotations, eliminating label noise inherent in manual datasets;
    \item We repurpose a pre-trained diffusion model as a multi-conditional proxy generator for dense pixel-to-surface correspondence prediction, enabling accurate and generalizable proxy estimation from synthetic training alone;
    \item We introduce hand refinement, which feeds enlarged hand crops as additional views for fine-grained hand recovery, and test-time scaling, which aggregates multiple stochastic predictions to estimate per-pixel uncertainty for robust fitting. Trained exclusively on synthetic data, DiffProxy achieves state-of-the-art results across five diverse real-world benchmarks.
\end{itemize}

%% file: sec/2_related_work.tex
\section{Related Work}
\label{sec:related_work}

\noindent\mbox{\textbf{End-to-end human mesh recovery.}}
Learning-based regression methods directly predict SMPL~\cite{SMPL:2015} or SMPL-X~\cite{SMPL-X:2019} parameters from images in a single forward pass. Early CNN-based approaches~\cite{kanazawa2019learning} have evolved into Transformer/ViT architectures~\cite{lin2021metro,goel2023humans4d,yin2025smplestx,chen2025human3r} trained on large-scale datasets~\cite{cheng2023dna,yi2023generating,lin2023motion,black2023bedlam}. These methods achieve fast inference but compressing geometry and correspondence reasoning into a single regression target tends to produce entangled errors, limiting achievable precision. Multi-view extensions~\cite{jia2023delving,li2024uhmr,zhu2025muc,matsubara2025heatformer} exploit geometric constraints but often suffer from limited training data and poor cross-dataset generalization. Recent work explores diffusion priors for HMR in parameter~\cite{stathopoulos2024score,cho2023generative,gao2025disrt}, mesh~\cite{foo2023distribution}, or video~\cite{zheng2025diffmesh,heo2024motion} space. In contrast, we generate dense correspondences as an intermediate representation for precise fitting-based recovery.

\noindent\mbox{\textbf{Fitting-based human mesh recovery.}}
Optimization-based methods fit parametric body models by minimizing objectives derived from detected cues. SMPLify~\cite{bogo2016smplify} and SMPLify-X~\cite{smplifyx} minimize 2D keypoint reprojection errors with pose priors and collision penalties. SPIN~\cite{kolotouros2019spin} combines regression initialization with iterative fitting. EasyMoCap~\cite{easymocap} extends this to multi-view settings with triangulated 3D keypoints. However, these methods depend on sparse keypoints that provide limited surface constraints and are sensitive to detection outliers. Our method replaces sparse keypoints with dense pixel-to-surface correspondences, providing far more constraints per view.

\noindent\mbox{\textbf{Dense correspondence for mesh recovery.}}
DensePose~\cite{guler2018densepose} pioneered pixel-to-surface correspondence for humans, inspiring \emph{iterative fitting} methods such as HoloPose~\cite{guler2019holopose} and DenseRaC~\cite{xu2019denserac} that optimize SMPL parameters against detected correspondences, and \emph{direct regression} methods like DecoMR~\cite{zeng2020decomr} and MeshPose~\cite{le2024meshpose} that generate meshes in a feed-forward manner. These approaches were hindered by noisy manual annotations in DensePose-COCO ($\sim$50K images) and the poor generalization of deterministic CNN detectors. Our work differs in three key aspects: (i)~synthetic training on 105K multi-view samples (844K images in total) with pixel-perfect annotations, eliminating label noise; (ii)~diffusion-based generation with strong visual priors for robust generalization; and (iii)~multi-view consistency via epipolar attention.

\noindent\mbox{\textbf{Diffusion models for dense prediction.}}
Pre-trained diffusion models have been successfully repurposed for dense prediction tasks. Marigold~\cite{ke2024marigold} and GenPercept~\cite{xu2024matters} demonstrated that Stable Diffusion~\cite{stabilityai2022sd21} backbones can be adapted for monocular depth and surface normal estimation while retaining zero-shot generalization. In parallel, enforcing multi-view consistency in diffusion models has been explored~\cite{shi2023mvdream,huang2024epidiff,yang2023consistnet,liu2024syncdreamer,jeong2024nvs,huang2024mvadapter,kant2024spad} for novel view synthesis and 3D generation, with SPAD~\cite{kant2024spad} introducing epipolar-constrained attention for cross-view interaction. We apply these principles to human dense correspondence prediction, leveraging diffusion priors to generate pixel-to-surface proxies for fitting.

\noindent\mbox{\textbf{Synthetic data for human mesh recovery.}}
Large-scale synthetic datasets~\cite{patel2021agora,yang2023synbody,black2023bedlam,yin2024whac} with precise SMPL-X annotations can approach or match real-data performance. We follow this paradigm with multi-view rendering, richer scene diversity (realistic occlusions, varied lighting, and clothing simulation), and pixel-perfect dense correspondence annotations, demonstrating zero-shot generalization to five diverse real-world benchmarks~\cite{mono-3dhp2017,wang20244ddress,Huang:CVPR:2022,bhatnagar22behave,tripathi2023ipman}.

%% file: sec/3_method.tex
\section{Method}
\label{sec:method}

Given multi-view images with calibrated cameras, our goal is to recover a precise SMPL-X body mesh. Rather than directly regressing model parameters from pixels, we train a diffusion-based generator on large-scale synthetic data with pixel-perfect annotations (\cref{subsec:data}) to predict dense pixel-to-surface correspondences from each input view (\cref{subsec:proxy}). SMPL-X parameters are then recovered by minimizing a unified reprojection objective against these proxies (\cref{subsec:fitting}).

\begin{figure*}[t]
\centering
\includegraphics[width=1.0\textwidth]{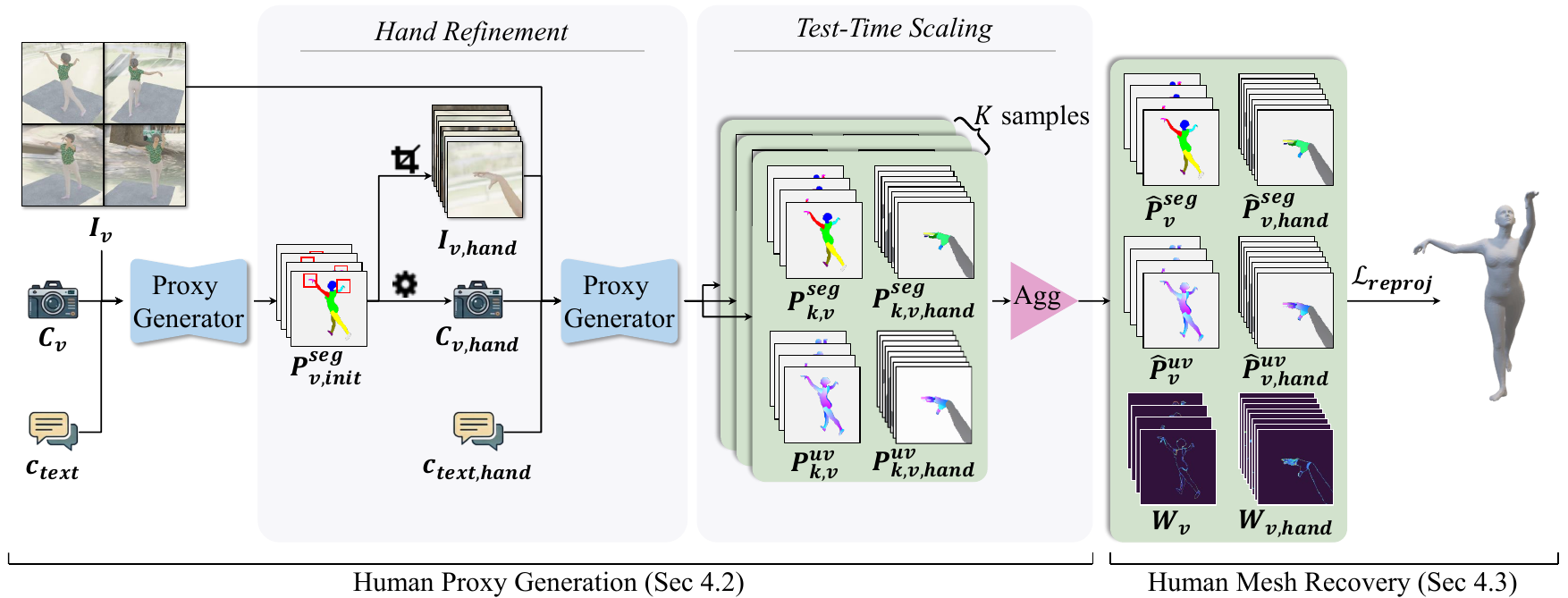}
\caption{Method overview. The proxy generator first produces initial body proxies; the segmentation component $\mathbf{P}_{v,\mathrm{init}}^{\mathrm{seg}}$ is then used to crop and enlarge hand regions for a second pass (hand refinement). Test-time scaling then draws $K$ stochastic samples for both body and hand proxies, aggregates them, and derives per-pixel uncertainty weight maps $\mathbf{W}_v$. Finally, SMPL-X parameters are recovered by minimizing an uncertainty-weighted reprojection objective $\mathcal{L}_{\mathrm{reproj}}$.}
\label{fig:pipeline_tts}
\end{figure*}

\begin{figure}[t]
    \centering
    \includegraphics[width=\linewidth]{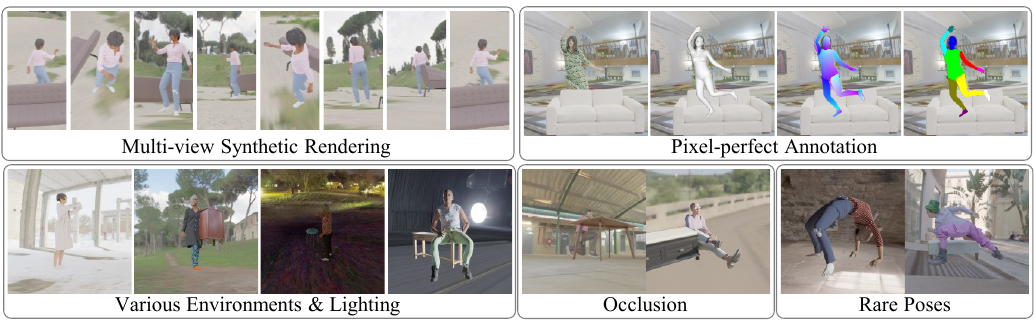}
    \caption{Synthetic training data samples. Our dataset covers diverse poses, realistic occlusions, varied lighting, and clothing simulation, all with pixel-perfect SMPL-X proxy annotations (segmentation and UV).}
    \label{fig:synth_data}
    \end{figure}
    
\subsection{Synthetic Data Preparation}
\label{subsec:data}

We train our model on a large-scale synthetic multi-view dataset with pixel-perfect SMPL-X annotations, eliminating annotation noise inherent in real-world datasets. A single model is trained on this dataset and used for all experiments without dataset-specific fine-tuning.

We construct our dataset by rendering clothed subjects from two sources: 67,650 from BEDLAM~\cite{black2023bedlam} driven by AMASS~\cite{AMASS:ICCV:2019} motion sequences, and 37,837 from SynBody~\cite{yang2023synbody} driven by MPI-3DHP~\cite{mono-3dhp2017} and MoYo~\cite{tripathi2023ipman} poses, totaling 105,487 unique SMPL-X subjects. We only use training-set motions from MPI-3DHP and MoYo; their test-set poses and all motions from our evaluation datasets are excluded. All visual content (cameras, backgrounds, textures, lighting) is entirely synthetic, ensuring that evaluation is unaffected by our data generation process. Our rendering pipeline incorporates realistic occlusions from 7,953 object meshes in Amazon Berkeley Objects~\cite{collins2022abo}, diverse hairstyles from Hair20K~\cite{he2025perm}, 863 HDR environment maps from Poly Haven~\cite{polyhaven} serving as both lighting and background, and physically-based clothing simulation~\cite{black2023bedlam}. For each subject, we sample 8 cameras and render 1024$\times$1024 RGB images with corresponding SMPL-X proxies, yielding 105,487 multi-view samples (843,896 images in total).

\begin{figure}[t]
    \centering
    \includegraphics[width=0.95\linewidth]{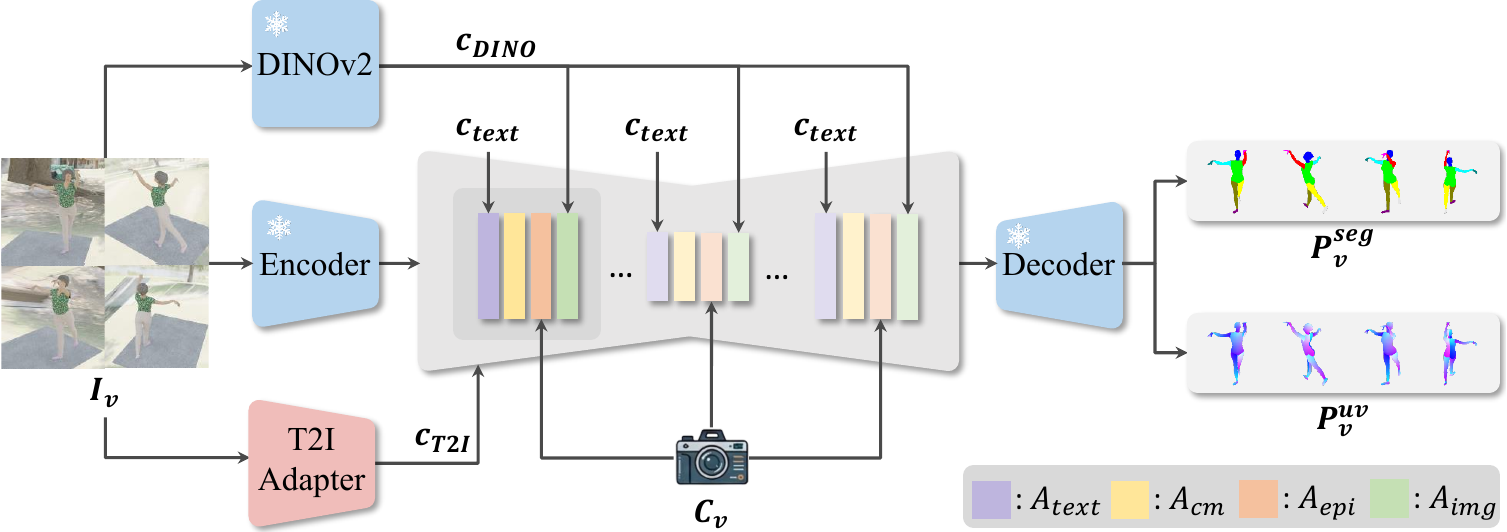}
    \caption{Diffusion-based proxy generator architecture. Our model is built on Stable Diffusion~2.1 with a frozen UNet backbone, equipped with three conditioning signals ($\mathbf{c}_{\text{text}}$, $\mathbf{c}_{\text{T2I}}$, $\mathbf{c}_{\text{DINO}}$) and four trainable attention modules ($\mathcal{A}_{\mathrm{text}}$, $\mathcal{A}_{\mathrm{img}}$, $\mathcal{A}_{\mathrm{cm}}$, $\mathcal{A}_{\mathrm{epi}}$) for multi-view consistent proxy generation.}
    \label{fig:architecture}
    \end{figure}
    
\subsection{Human Proxy Generation}
\label{subsec:proxy}

\noindent\mbox{\textbf{SMPL-X and proxy definition.}}
SMPL-X~\cite{SMPL-X:2019} is a parametric 3D human mesh model with parameters $\Theta = \{\boldsymbol{\beta}, \boldsymbol{\theta}, \boldsymbol{\psi}, \mathbf{T}\}$: shape $\boldsymbol{\beta}\!\in\!\mathbb{R}^{10}$, pose $\boldsymbol{\theta}$, facial expression $\boldsymbol{\psi}$, and global translation $\mathbf{T}\!\in\!\mathbb{R}^3$. Each vertex carries a 2D uv coordinate $\mathbf{u}\!\in\![0,1]^2$ on a predefined texture map partitioned by body parts.

We define SMPL-X proxy as a 2D dense representation establishing pixel-to-surface correspondences. For view $v$, the proxy $\mathbf{P}_v = (\mathbf{P}_v^{\mathrm{seg}}, \mathbf{P}_v^{\mathrm{uv}})$ consists of segmentation and UV components. Both are encoded as three-channel images to match the pre-trained VAE decoder of SD~2.1, which empirically yields better reconstruction quality than single-channel alternatives. $\mathbf{P}_v^{\mathrm{seg}}$ assigns each semantic body part a unique RGB color. $\mathbf{P}_v^{\mathrm{uv}}$ stores the uv coordinates in the first two channels with the third set to one. We render the textured mesh with perspective projection to obtain the proxy images.

\noindent\mbox{\textbf{Diffusion-based proxy generator.}}
Our generator $G_\phi$ is built on Stable Diffusion~2.1~\cite{stabilityai2022sd21} with frozen UNet backbone. Given multi-view images $\{I_v\}_{v=1}^{N}$ ($N\geq 1$) and perspective camera parameters $\{C_v\}=\{(\mathbf{K}_v,\mathbf{R}_v,\mathbf{t}_v)\}$ (no lens distortion), the model predicts SMPL-X proxies $\mathbf{P}_v = (\mathbf{P}_v^{\mathrm{seg}},\mathbf{P}_v^{\mathrm{uv}}) \!\in\! \mathbb{R}^{256\times 256\times 3}$ encoding body part labels and surface coordinates.

Three conditioning signals drive the generation: text prompts $\mathbf{c}_{\text{text}}$ control output modality; T2I-Adapter~\cite{mou2023t2i} features $\mathbf{c}_{\text{T2I}} = \mathcal{E}_{\text{T2I}}(I_v)$ enforce pixel-level alignment; DINOv2~\cite{oquab2023dinov2} tokens $\mathbf{c}_{\text{DINO}} = \mathcal{E}_{\text{DINO}}(I_v)$ provide pose and appearance priors. $\mathbf{c}_{\text{text}}$ and $\mathbf{c}_{\text{DINO}}$ are injected via text cross-attention $\mathcal{A}_{\text{text}}$ and image cross-attention $\mathcal{A}_{\text{img}}$, while $\mathbf{c}_{\text{T2I}}$ is added as residual features. For multi-modal and multi-view consistency, trainable cross-modality attention $\mathcal{A}_{\text{cm}}$ concatenates UV and segmentation tokens, and multi-view epipolar attention $\mathcal{A}_{\text{epi}}$ enforces geometric consistency via epipolar-constrained self-attention~\cite{kant2024spad} with Plücker ray embeddings. Only the attention modules and T2I-Adapter are trained; the UNet and DINOv2 remain frozen.

The model is trained with the standard diffusion objective. Given ground-truth proxy $\mathbf{P}_v^*$, we encode it to latent $\mathbf{z}_0 = \mathcal{E}_{\text{VAE}}(\mathbf{P}_v^*)$ via the VAE encoder, sample timestep $t$ and noise $\boldsymbol{\epsilon} \sim \mathcal{N}(\mathbf{0}, \mathbf{I})$, and optimize:
\begin{equation}
\mathcal{L}_{\text{diff}} = \mathbb{E}_{\mathbf{z}_0, \boldsymbol{\epsilon}, t, \mathbf{c}} \left[\|\boldsymbol{\epsilon} - \boldsymbol{\epsilon}_\phi(\mathbf{z}_t, t, \mathbf{c})\|_2^2\right],
\label{eq:diffusion_loss}
\end{equation}
where $\mathbf{z}_t = \sqrt{\bar{\alpha}_t}\mathbf{z}_0 + \sqrt{1-\bar{\alpha}_t}\boldsymbol{\epsilon}$ and $\mathbf{c} = \{\mathbf{c}_{\text{text}}, \mathbf{c}_{\text{T2I}}, \mathbf{c}_{\text{DINO}}\}$. We train with a fixed budget of $N\!=\!4$ views per sample, where 2–4 views are full-body and the remaining slots are filled with hand-region crops (left/right randomly selected) from the same camera viewpoints. This enables joint body and hand training without architectural changes. At inference, we denoise a random latent $\mathbf{z}_T \sim \mathcal{N}(\mathbf{0}, \mathbf{I})$ and decode via VAE decoder $\mathcal{D}$ to obtain $\mathbf{P}_v$. The model generalizes to different view counts without fine-tuning.

\noindent\mbox{\textbf{Hand refinement.}}
In full-body images $I_v$, hands occupy $<$1\% of pixels, producing sparse and unreliable proxy predictions in these regions. We adopt a two-pass strategy: first inferring initial body proxies $\mathbf{P}_{v,\mathrm{init}}$, then using the segmentation component $\mathbf{P}_{v,\mathrm{init}}^{\mathrm{seg}}$ to localize hand regions and create enlarged crops $I_{v,\mathrm{hand}}$. When cropping, extrinsic parameters remain unchanged while intrinsics (focal length and principal point) are adjusted to produce hand-specific camera parameters $C_{v,\mathrm{hand}}$. In the second pass, we treat hand crops as additional views with $C_{v,\mathrm{hand}}$ and switch to a hand-specific text prompt $\mathbf{c}_{\text{text,hand}}$ that subdivides each hand into 12 parts (two palms and ten fingers), leveraging cross-view attention to produce refined hand proxies $\mathbf{P}_{v,\mathrm{hand}}$. This coarse-to-fine strategy improves finger fidelity without modifying the network architecture.

\noindent\mbox{\textbf{Test-time scaling \& uncertainty.}}
Individual pixel predictions can still be erroneous in challenging regions such as self-occlusions, part boundaries, or visually ambiguous areas. We leverage diffusion stochasticity to identify and mitigate such errors: by drawing $K$ stochastic samples $\{\mathbf{P}_{k,v}\}_{k=1}^{K}$ per view and measuring their disagreement, we obtain per-pixel uncertainty estimates that down-weight unreliable predictions during optimization.

For UV aggregation, we compute the pixel-wise median across samples to obtain a robust estimate:
\begin{equation}
\hat{\mathbf{P}}_v^{\mathrm{uv}}(x) \;=\; \operatorname{median}_{k=1..K}\!\big[\,\mathbf{P}_{k,v}^{\mathrm{uv}}(x)\,\big],
\label{eq:uv_median}
\end{equation}
and quantify uncertainty using the channel-wise sample variance averaged over three channels:
\begin{equation}
\mathbf{U}_v^{\mathrm{uv}}(x) \;=\; \frac{1}{3}\sum_{c=1}^{3}\operatorname{Var}_{k=1..K}\!\big[\,\mathbf{P}_{k,v}^{\mathrm{uv},(c)}(x)\,\big].
\label{eq:uv_uncertainty}
\end{equation}

For segmentation, we first quantize each sample $\mathbf{P}_{k,v}^{\mathrm{seg}}$ to the nearest color in a predefined palette $\mathcal{P}_{\mathrm{view}}$ (hand part palette for hand crops, body part palette otherwise). We then apply pixel-wise majority voting across $K$ samples to obtain the aggregated segmentation $\hat{\mathbf{P}}_v^{\mathrm{seg}}$. The segmentation uncertainty measures how strongly the $K$ samples agree: if the most-voted label at pixel $x$ receives $n_{\max}(x)$ votes, the uncertainty is
\begin{equation}
\mathbf{U}_v^{\mathrm{seg}}(x) \;=\;
\begin{cases}
1, & n_{\max}(x)\le \tfrac{K}{2},\\[2pt]
2\!\left(1-\dfrac{n_{\max}(x)}{K}\right), & \text{otherwise},
\end{cases}
\label{eq:seg_uncertainty}
\end{equation}
which equals 1 (maximum) when no label achieves majority and decreases linearly to 0 as agreement reaches 100\%.

The uncertainties modulate the fitting via a per-view weight map $\mathbf{W}_v \!\in\! \mathbb{R}^{256\times 256}$, where each pixel $x$ is assigned a reliability weight:
\begin{equation}
\mathbf{W}_v(x) \;=\; \big(1-\mathbf{U}_v^{\mathrm{uv}}(x)\big)\,\big(1-\mathbf{U}_v^{\mathrm{seg}}(x)\big).
\label{eq:pixel_weight}
\end{equation}
This strategy provides a compute--accuracy trade-off through $K$ without test-time adaptation of network weights.

\subsection{Human Mesh Recovery}
\label{subsec:fitting}

Our dense proxies provide far richer constraints than sparse keypoints: each foreground pixel carries a semantic part label and UV coordinate on the SMPL-X surface, offering dense coverage across the entire visible body.

Given proxies $\{\mathbf{P}_v\}$ from all views, we compute the reprojection loss over all foreground pixels. Let $\text{fg}(v)$ denote foreground pixels in view $v$. For each pixel $x\!\in\!\text{fg}(v)$, we map it to a 3D point on the SMPL-X surface via its proxy label and UV coordinate, and compute the L2 reprojection error $d(x)$ between the projected surface point and the original pixel (see Algorithm~1 in supplementary). The reprojection loss, weighted by the per-pixel reliability map $\mathbf{W}_v$ (\cref{eq:pixel_weight}), is:
\begin{equation}
\mathcal{L}_{\text{reproj}} = \sum_{v}\sum_{x\in\text{fg}(v)} \mathbf{W}_v(x)\, d(x)^2.
\label{eq:reproj_loss}
\end{equation}
When test-time scaling is disabled ($K\!=\!1$), all weights default to $\mathbf{W}_v(x)\!=\!1$.

\noindent\mbox{\textbf{Optimization.}}
We optimize body and hand poses in axis-angle space, and shape $\boldsymbol{\beta}$ without explicit regularization. Body and hand poses are regularized by DPoser-X~\cite{lu2025dposerx}, a diffusion-based whole-body pose prior, via a one-step denoising loss $\mathcal{L}_{\text{prior}}$. The total objective is:
\begin{equation}
\mathcal{L} = \mathcal{L}_{\text{reproj}} + \lambda_{\text{prior}}\, \mathcal{L}_{\text{prior}}, \quad \lambda_{\text{prior}} = 0.1.
\label{eq:total_loss}
\end{equation}
We minimize $\mathcal{L}$ using Adam with stage-wise learning rate scheduling. See \cref{sec:experiments} for implementation details.

%% file: sec/4_experiments.tex
\section{Experiments}
\label{sec:experiments}

\subsection{Implementation Details}

\noindent\mbox{\textbf{Proxy generator training.}}
We trained with 4 views per sample using random full-body/hand crops and bbox augmentation. From SD-2.1 pre-trained weights, we optimized only the attention modules ($\mathcal{A}_{\text{text}}$, $\mathcal{A}_{\text{img}}$, $\mathcal{A}_{\text{cm}}$, $\mathcal{A}_{\text{epi}}$) and T2I-Adapter $\mathcal{E}_{\text{T2I}}$ while freezing the UNet backbone and DINOv2 $\mathcal{E}_{\text{DINO}}$. Training used batch size 2, Adam optimizer, learning rate $5\times 10^{-5}$, for 30 epochs on 4$\times$ RTX 5090 GPUs ($\sim$36 hours).

\noindent\mbox{\textbf{VAE decoder refinement.}}
Stable Diffusion's pre-trained VAE decoder $\mathcal{D}$ may introduce quantization artifacts for proxy representations that require high numerical precision. We fine-tuned $\mathcal{D}$ with learning rate $1\times 10^{-6}$, batch size 8, for 100K iterations ($\sim$4 hours on 4$\times$ RTX 5090).

\noindent\mbox{\textbf{Inference.}}
We generate proxies $\mathbf{P}_v$ for 12 views by default: 4 full-body views plus left/right hand crops for each. Inference involves two passes on a single RTX 5090: first generating 4 full-body proxies to localize hand regions ($\sim$3s), then generating all 12 proxies including hand crops ($\sim$15s). With test-time scaling, proxy generation takes $K \times 15$s (default $K=3$). Mesh fitting adds $\sim$50s without hand refinement or $\sim$60s with it.

\noindent\mbox{\textbf{Mesh fitting.}}
We optimize $\boldsymbol{\beta}$, $\boldsymbol{\theta}$, and $\mathbf{T}$ from $\Theta$, along with a global scale parameter, using Adam in a coarse-to-fine schedule that first solves global placement, then body pose and shape, and finally hand articulations. We advance to the next stage when the loss converges, \ie, the relative decrease between consecutive iterations falls below a per-stage threshold (1\%--10\%). Facial expression $\boldsymbol{\psi}$ is not optimized as our focus is on body and hand reconstruction.

\subsection{Datasets and Baselines}

\noindent\mbox{\textbf{Datasets.}}
We evaluate on five real-world datasets: 3DHP~\cite{mono-3dhp2017}, BEHAVE~\cite{bhatnagar22behave}, RICH~\cite{Huang:CVPR:2022}, MoYo~\cite{tripathi2023ipman}, and 4D-DRESS~\cite{wang20244ddress}, covering studio capture, human-object interaction, outdoor scenes, challenging poses, and loose clothing. We additionally test on 4D-DRESS with random crops (4D-DRESS partial) to evaluate robustness to partial observations. All real datasets are unseen during training; a single model is used without dataset-specific fine-tuning.

\noindent\mbox{\textbf{Baselines.}}
We compare against six baselines spanning three categories: \emph{single-view end-to-end}---SMPLest-X~\cite{yin2025smplestx} (trained on large-scale diverse data) and Human3R~\cite{chen2025human3r} (extending CUT3R for joint human-scene recovery); \emph{multi-view end-to-end}---U-HMR~\cite{li2024uhmr} (decoupled camera and body estimation), MUC~\cite{zhu2025muc} (prediction fusion), and HeatFormer~\cite{matsubara2025heatformer} (neural optimization with heatmaps); and \emph{fitting-based}---EasyMoCap~\cite{easymocap} (sparse keypoint fitting).

\subsection{Quantitative Results}

\noindent\mbox{\textbf{Evaluation protocol.}}
We evaluate on official test/validation splits with a fixed set of 4 views per dataset. On 3DHP~\cite{mono-3dhp2017}, we follow HeatFormer~\cite{matsubara2025heatformer} and use subject S8 with views 0/2/7/8, filtering invalid masks as in their protocol. BEHAVE~\cite{bhatnagar22behave} uses the official test split with all 4 views; RICH~\cite{Huang:CVPR:2022} uses the first 4 valid views from the test split; MoYo~\cite{tripathi2023ipman} uses the validation split with views 1/3/4/5. Since 4D-DRESS~\cite{wang20244ddress} lacks an official split, we evaluate on the full dataset with all 4 views. To reduce temporal redundancy, we sample every 5th frame~\cite{rogez2019lcr,black2023bedlam,lassner2017unite}; for 4D-DRESS we sample every 50th frame to keep evaluation tractable. We report MPJPE, MPVPE, and their Procrustes-aligned variants in millimeters. For single-view baselines, we evaluate each camera view independently after root alignment and report the single view with the lowest MPVPE per dataset, giving these methods a favorable comparison.

\begin{table*}[t]
  \centering
  \caption{Quantitative comparison on five real-world datasets. $^*$ indicates the method was trained on that specific dataset. For single-view methods, the best single-camera result is reported. Three configurations: ``w/o HR, TTS'' (4 full-body views only), ``w/o TTS'' (+ hand refinement), ``Ours'' (full model). \colorbox{HeatBest}{Best} / \colorbox{HeatSecond}{second best}.}
  \label{tab:main_results}%
  \resizebox{\textwidth}{!}{
  \begin{tabular}{l|cccc|cccc|cccc}
  \hline
  & \multicolumn{4}{c|}{\textit{3dhp}} & \multicolumn{4}{c|}{\textit{rich}} & \multicolumn{4}{c}{\textit{behave}} \\
  \hline
  Method & PA-MPJPE & MPJPE & PA-MPVPE & MPVPE & PA-MPJPE & MPJPE & PA-MPVPE & MPVPE & PA-MPJPE & MPJPE & PA-MPVPE & MPVPE \\
  \hline
  SMPLest-X~\cite{yin2025smplestx} & \second{32.5}$^*$ & 48.1$^*$ & 46.6$^*$ & 62.3$^*$ & 31.8$^*$ & 67.6$^*$ & 40.3$^*$ & 81.1$^*$ & 30.0$^*$ & 48.6$^*$ & 43.5$^*$ & 63.6$^*$ \\
  Human3R~\cite{chen2025human3r} & 53.1 & 84.3 & 67.7 & 104.1 & 52.1 & 101.1 & 64.4 & 116.9 & 38.3 & 85.0 & 53.8 & 102.3 \\
  \hline
  U-HMR~\cite{li2024uhmr} & 67.3$^*$ & 142.9$^*$ & 78.5$^*$ & 164.2$^*$ & 55.5 & 134.9 & 67.8 & 159.3 & 46.1 & 118.7 & 51.9 & 135.0 \\
  MUC~\cite{zhu2025muc} & 40.2 & - & 51.3 & - & 35.7$^*$ & - & 42.7$^*$ & - & 28.7 & - & 41.8 & - \\
  HeatFormer~\cite{matsubara2025heatformer} & \best{29.2}$^*$ & 49.9$^*$ & \best{34.7}$^*$ & 53.3$^*$ & 40.1 & 80.1 & 52.9 & 93.8 & 33.4 & 70.2 & 47.7 & 80.5 \\
  EasyMoCap~\cite{easymocap} & 34.3 & 43.7 & \second{43.2} & 51.5 & 31.2 & 49.8 & 44.2 & 60.3 & \best{21.8} & \best{29.6} & 35.1 & \second{40.4} \\
  \hline
  Ours (w/o HR, TTS) & 33.4 & 42.0 & 43.4 & \best{50.5} & 25.6 & 36.6 & 37.5 & 46.5 & 23.6 & 33.7 & 33.7 & 43.1 \\
  Ours (w/o TTS) & 33.1 & \second{41.4} & 45.3 & 51.5 & \second{24.5} & \second{34.6} & \second{29.3} & \second{36.2} & 23.5 & 33.5 & \second{32.2} & 40.7 \\
  Ours & 32.8 & \best{41.1} & 44.7 & \second{50.9} & \best{24.2} & \best{33.7} & \best{28.9} & \best{35.1} & \second{22.8} & \second{32.7} & \best{31.3} & \best{39.9} \\
  \hline
  \hline
  & \multicolumn{4}{c|}{\textit{moyo}} & \multicolumn{4}{c|}{\textit{4ddress}} & \multicolumn{4}{c}{\textit{4ddress-partial}} \\
  \hline
  Method & PA-MPJPE & MPJPE & PA-MPVPE & MPVPE & PA-MPJPE & MPJPE & PA-MPVPE & MPVPE & PA-MPJPE & MPJPE & PA-MPVPE & MPVPE \\
  \hline
  SMPLest-X~\cite{yin2025smplestx} & 45.6$^*$ & 63.3$^*$ & 62.1$^*$ & 81.4$^*$ & 34.8 & 52.1 & 52.4 & 70.2 & 72.2 & 101.7 & 111.3 & 140.6 \\
  Human3R~\cite{chen2025human3r} & 65.0 & 87.8 & 81.6 & 107.6 & 30.0 & 52.4 & 42.4 & 66.7 & 72.3 & 117.6 & 103.5 & 153.9 \\
  \hline
  U-HMR~\cite{li2024uhmr} & 88.3 & 177.7 & 104.9 & 209.7 & 42.1 & 80.3 & 52.6 & 98.8 & 65.2 & 136.1 & 87.2 & 175.2 \\
  MUC~\cite{zhu2025muc} & 62.8 & - & 76.0 & - & 33.2 & - & 49.2 & - & 60.1 & - & 92.9 & - \\
  HeatFormer~\cite{matsubara2025heatformer} & 67.3 & 126.9 & 79.6 & 123.6 & 43.4 & 73.3 & 61.6 & 90.2 & 143.8 & 291.6 & 175.2 & 327.0 \\
  EasyMoCap~\cite{easymocap} & 31.4 & 41.2 & 44.8 & 51.6 & 19.9 & 24.9 & 31.8 & 35.3 & 48.2 & 94.8 & 69.8 & 119.4 \\
  \hline
  Ours (w/o HR, TTS) & 29.4 & 33.6 & 41.8 & 44.9 & 17.6 & 22.4 & 30.4 & 33.4 & 29.6 & 34.4 & 49.6 & 52.0 \\
  Ours (w/o TTS) & \second{29.2} & \second{33.0} & \second{36.3} & \second{39.6} & \second{17.0} & \second{20.8} & \second{24.0} & \second{26.2} & \second{28.1} & \second{32.6} & \second{42.3} & \second{44.4} \\
  Ours & \best{28.3} & \best{32.4} & \best{35.2} & \best{38.8} & \best{16.6} & \best{20.5} & \best{23.3} & \best{25.6} & \best{27.6} & \best{31.6} & \best{41.3} & \best{43.1} \\
  \hline
  \end{tabular}}
\end{table*}%

As shown in \cref{tab:main_results}, our method achieves state-of-the-art performance across all six evaluation settings. We organize our analysis around three observations:

\textbf{Fitting-based vs.\ end-to-end methods.} End-to-end methods generally exhibit higher errors than fitting-based approaches on absolute positioning metrics. For example, on RICH the best end-to-end MPVPE is 81.1\,mm (SMPLest-X), whereas EasyMoCap achieves 60.3\,mm and our full model 35.1\,mm, confirming that constructing accurate intermediate representations and fitting against them is more effective than direct regression. Notably, even our base model (w/o HR, TTS)---using the same 4-view input as multi-view baselines---already surpasses all end-to-end methods on most metrics.

\textbf{Cross-dataset generalization.} Existing multi-view methods are limited by scarce real multi-view data and tend to excel only on their training domains---\eg, HeatFormer~\cite{matsubara2025heatformer} achieves competitive MPVPE (53.3) on 3DHP where it is trained, but drops to 123.6 on MoYo and 327.0 on 4D-DRESS partial, falling behind even single-view methods. Our large-scale synthetic dataset (105K multi-view samples) combines data diversity with multi-view geometric supervision, enabling strong performance across all benchmarks without real training data.

\textbf{Dense proxies vs.\ sparse keypoints.} EasyMoCap~\cite{easymocap} is also fitting-based but relies on sparse keypoints. The advantage of dense proxies manifests in two aspects. (1)~\emph{Precision}: our proxy maps each foreground pixel to a specific surface point, constraining the full mesh surface rather than just joint locations. We outperform EasyMoCap in MPVPE on all six settings (\eg, 35.1 vs.\ 60.3 on RICH, 25.6 vs.\ 35.3 on 4D-DRESS). (2)~\emph{Robustness}: EasyMoCap performs well where keypoint detection is reliable (3DHP, BEHAVE) but degrades under complex lighting (RICH), extreme poses (MoYo), and loose clothing (4D-DRESS). The gap widens with partial views: EasyMoCap's MPVPE increases by 238\% from 4D-DRESS to 4D-DRESS partial (35.3$\to$119.4), while ours increases by only 68\% (25.6$\to$43.1), as dense constraints across the visible surface make the fitting resilient to partial occlusion.

\begin{figure*}[t]
\centering
\includegraphics[width=\linewidth]{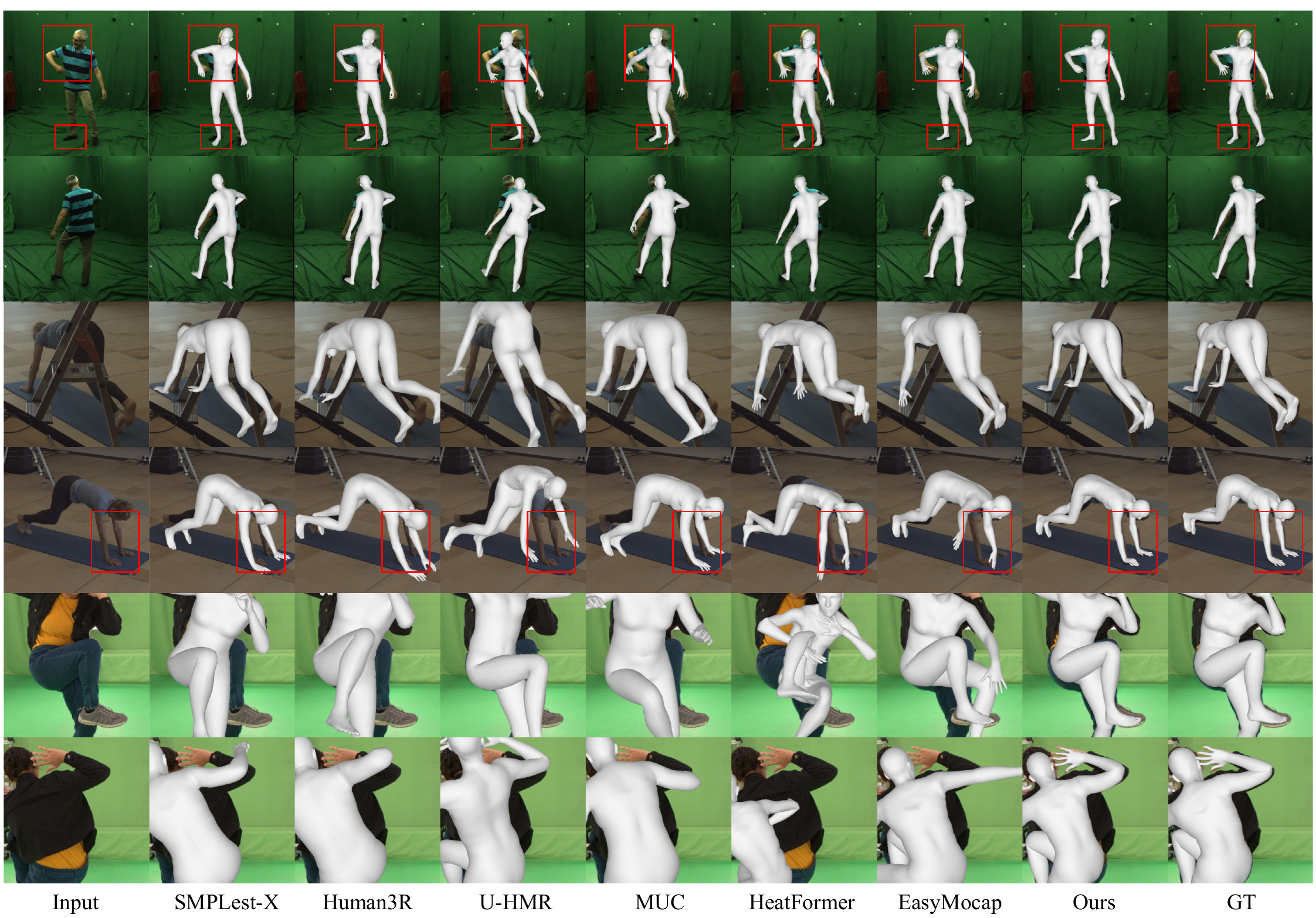}
\caption{Qualitative comparison across three scenarios: 3DHP (rows~1--2), RICH (rows~3--4), and 4D-DRESS partial (rows~5--6). Red boxes highlight notable misalignment. Our method maintains tight image-mesh correspondence across all settings.}
\label{fig:qualitative}
\end{figure*}

\subsection{Qualitative Results}

\Cref{fig:qualitative} presents qualitative comparisons across three representative scenarios. On 3DHP (rows~1--2), the ground-truth meshes show visible offset from the images in certain regions (red boxes in row~1); methods trained on this dataset (HeatFormer, SMPLest-X) tend to reproduce similar alignment patterns, suggesting that their metrics partly reflect annotation biases. Our method achieves tighter visual alignment despite lower scores on this benchmark. On RICH (rows~3--4), the cluttered real-world environment misleads EasyMoCap's sparse keypoint detection, resulting in degraded fitting quality, while our dense proxies remain robust to such visual distractors. On 4D-DRESS partial (rows~5--6), most methods degrade severely with limited visibility, whereas our dense surface constraints remain effective even under heavy occlusion.

\subsection{Ablation Studies}

\begin{figure}[t]
\centering
\includegraphics[width=\linewidth]{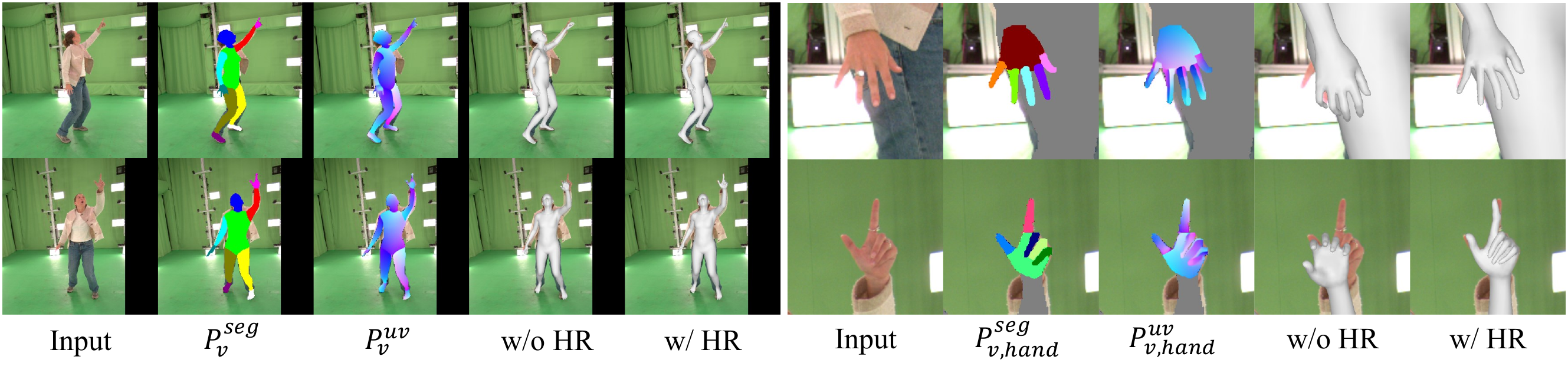}
\caption{Qualitative comparison of hand refinement. Hand refinement improves fitting quality and produces accurate finger details.}
\label{fig:hand_refine_qual}
\end{figure}

\noindent\mbox{\textbf{Hand refinement.}}
Comparing ``w/o HR, TTS'' and ``w/o TTS'' in \cref{tab:main_results} reveals that hand refinement consistently improves accuracy. In full-body images, hands occupy $<$1\% of pixels; hand refinement produces clearer proxies that better localize hand position, anchoring arm orientation and improving full-body accuracy. \Cref{fig:hand_refine_qual} shows that refinement produces hand proxies with more accurate finger poses and less part ambiguity.

\begin{figure}[t]
  \centering
  \includegraphics[width=\linewidth]{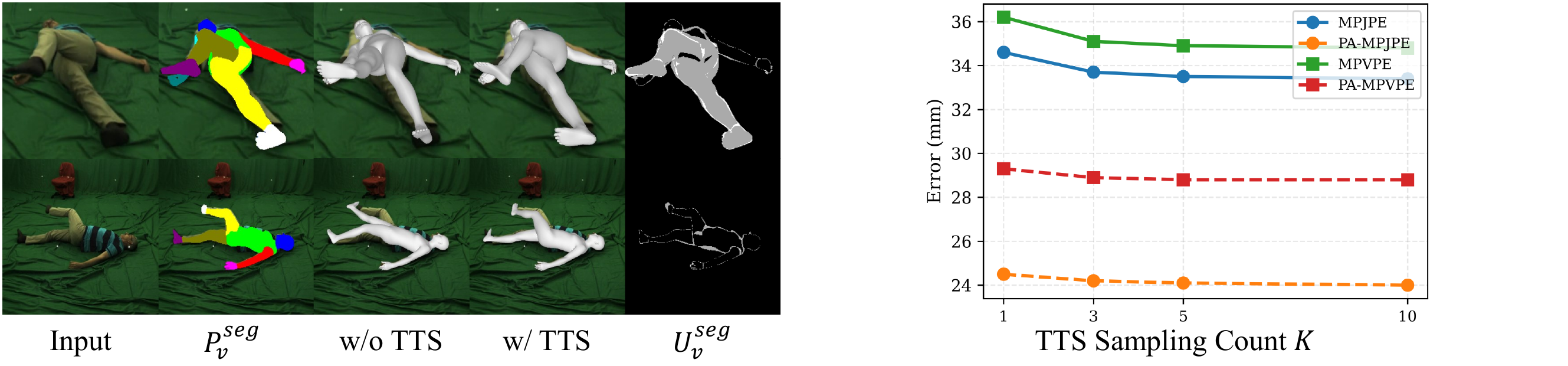}
  \caption{Test-time scaling with uncertainty weighting improves robustness by down-weighting unreliable predictions and recovering correct poses from erroneous proxy outputs (left). Increasing $K$ consistently improves reconstruction quality (right).}
  \label{fig:tts_ablation}
  \end{figure}

\noindent\mbox{\textbf{Test-time scaling and uncertainty weighting.}}
\Cref{fig:tts_ablation} illustrates: a single prediction may swap the left and right leg labels, but across $K$ stochastic samples some predictions are correct while others are not; this disagreement is captured by $\mathbf{U}_v^{\mathrm{seg}}$, which assigns high uncertainty to inconsistent regions. Fitting then down-weights these pixels and relies on more consistent views to recover the correct configuration. Increasing $K$ generally improves performance (\cref{fig:tts_ablation}, right); we use $K\!=\!3$ as default.

\begin{figure}[t]
  \centering
  \includegraphics[width=\linewidth]{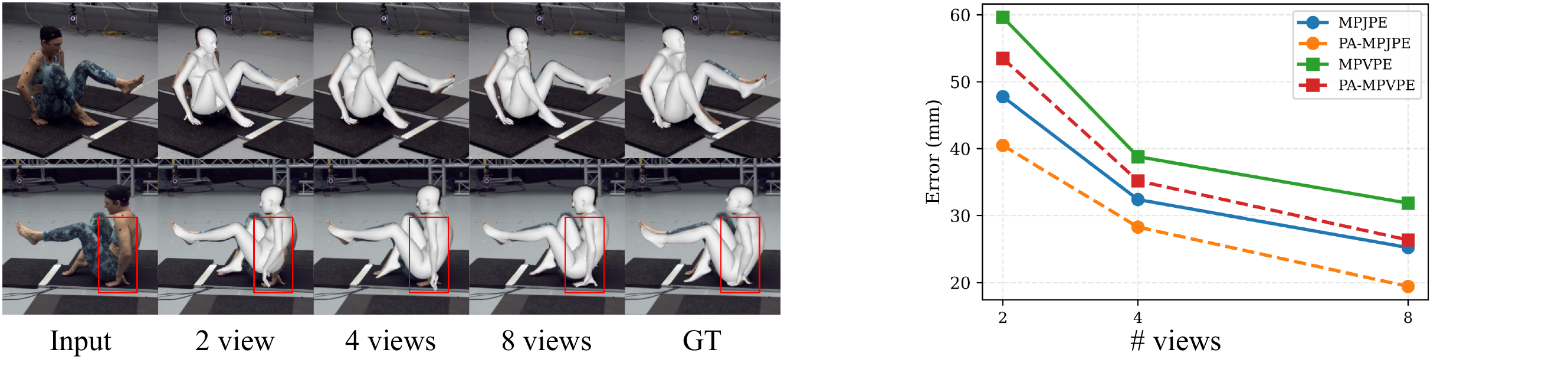}
  \caption{Our method benefits from increasing view counts, with performance improving progressively from 2-view to 8-view configurations.}
  \label{fig:view_ablation}
  \end{figure}

\noindent\mbox{\textbf{Number of input views.}}
Our method supports flexible view counts without retraining. As shown in \cref{fig:view_ablation} (evaluated on MoYo), even two views already produce reasonable reconstructions. Adding more views further improves the recovery of occluded details---\eg, arms behind the torso and hands near the hip (red boxes) become progressively more accurate as additional viewpoints increase their visibility.

\begin{table}[t]
  \centering
  \caption{Ablation study on BEHAVE. Top: removing individual network modules. Bottom: inference without ground-truth cameras.}
  \label{tab:module_ablation}
  \label{tab:camera_ablation}
  \scalebox{0.8}{
  \begin{tabular}{l|cccc}
  \hline
  Configuration & PA-MPJPE & MPJPE & PA-MPVPE & MPVPE \\
  \hline
  w/o DINOv2 & 37.3 & 44.3 & 53.7 & 60.2 \\
  w/o T2I-Adapter & 28.5 & 41.5 & 36.7 & 48.4 \\
  w/o $\mathcal{A}_{\text{text}}$ & 26.5 & 35.2 & 35.3 & 43.3 \\
  w/o $\mathcal{A}_{\text{epi}}$ & 25.7 & 33.7 & 35.0 & 41.9 \\
  w/o $\mathcal{A}_{\text{cm}}$ & 26.3 & 33.8 & 35.4 & 41.9 \\
  \hline
  w/o GT camera & 25.9 & -- & 37.3 & -- \\
  \hline
  Ours & \textbf{22.8} & \textbf{32.7} & \textbf{31.3} & \textbf{39.9} \\
  \hline
  \end{tabular}}
\end{table}

\noindent\mbox{\textbf{Network module contributions.}}
\Cref{tab:module_ablation} ablates individual network modules on BEHAVE. We independently remove DINOv2, T2I-Adapter, and attention modules ($\mathcal{A}_{\text{text}}$, $\mathcal{A}_{\text{epi}}$, $\mathcal{A}_{\text{cm}}$). Each module contributes to overall performance, with the full model achieving the best results.

\noindent\mbox{\textbf{Inference without camera calibration.}}
While our main results assume calibrated cameras, real-world scenarios often lack ground-truth camera parameters. We test a camera-free variant on BEHAVE: we predict camera parameters using VGGT~\cite{wang2025vggt}, then generate proxies with these predicted cameras. During mesh fitting, we jointly optimize camera parameters alongside body pose and shape to compensate for prediction inaccuracy. As shown in \cref{tab:camera_ablation}, our method achieves competitive performance with only moderate degradation.

\begin{figure*}[t]
\centering
\begin{minipage}[t]{0.49\linewidth}
\centering
\includegraphics[width=\linewidth]{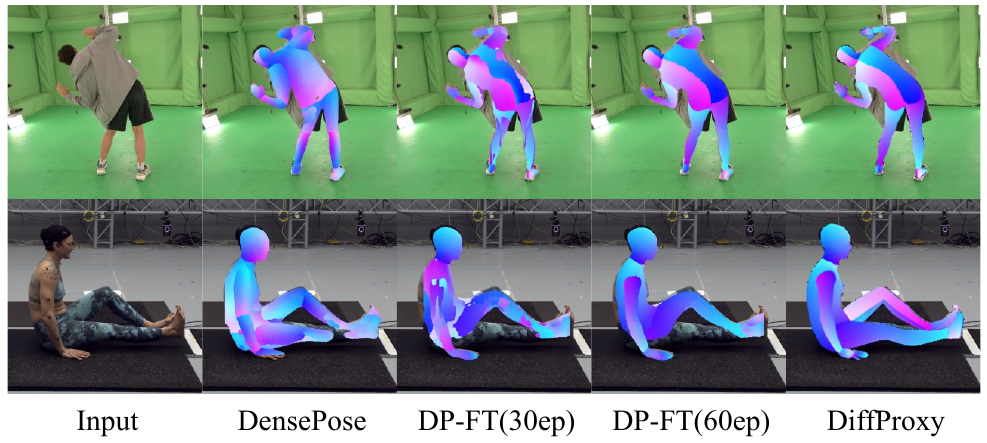}
\caption{Comparison with DensePose~\cite{guler2018densepose} and its fine-tuned variants (DP-FT). Fine-tuning improves standard poses (left) but provides limited benefit on challenging ones (right). DiffProxy produces accurate proxies in both cases.}
\label{fig:densepose_ablation}
\end{minipage}
\hfill
\begin{minipage}[t]{0.49\linewidth}
\centering
\includegraphics[width=\linewidth]{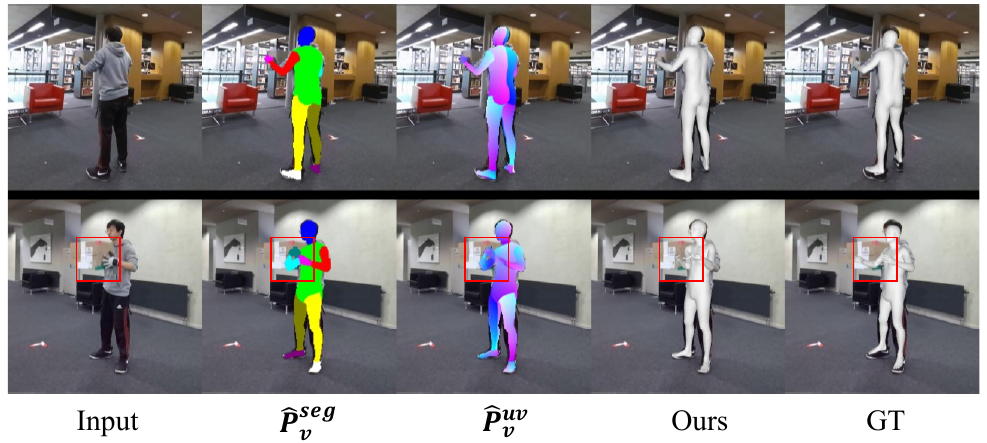}
\caption{A typical failure case on BEHAVE. The proxy predicts correct dense correspondences, but mesh fitting converges to a local optimum with inverted joint orientations---the pose appears correct in 2D but is flipped in depth.}
\label{fig:failed_case}
\end{minipage}
\end{figure*}

\noindent\mbox{\textbf{Why diffusion-based proxy generation?}}
A natural alternative is a discriminative model such as DensePose~\cite{guler2018densepose}. We fine-tune DensePose on our synthetic dataset for 30 and 60 epochs and run our fitting pipeline on its predictions. \Cref{fig:densepose_ablation} shows qualitative results; \cref{tab:densepose_ablation} reports the better variant (60 epochs). Fine-tuning improves DensePose on standard poses, confirming that our synthetic data provides effective supervision, but even with extended training DensePose still produces fragmented predictions on challenging poses. Our diffusion-based generator leverages the visual priors of Stable Diffusion~2.1 for robust generalization across diverse image compositions, naturally supporting hand refinement on enlarged crops, and enabling test-time scaling for uncertainty estimation via stochastic sampling.

\begin{table}[t]
  \centering
  \caption{Comparison with DensePose~\cite{guler2018densepose} and its fine-tuned variant on MoYo.}
  \label{tab:densepose_ablation}
  \scalebox{0.8}{
  \begin{tabular}{l|cccc}
  \hline
  Method & PA-MPJPE & MPJPE & PA-MPVPE & MPVPE \\
  \hline
  DensePose & 67.6 & 101.1 & 90.4 & 127.8 \\
  DP-FT (60ep) & 66.5 & 88.8 & 84.5 & 111.4 \\
  \hline
  Ours & \textbf{28.3} & \textbf{32.4} & \textbf{35.2} & \textbf{38.8} \\
  \hline
  \end{tabular}}
\end{table}

%% file: sec/5_limitation_and_future_works.tex
\section{Limitation and Future Works}
\label{sec:limitation_and_future_works}

While DiffProxy achieves state-of-the-art performance, several limitations remain. \textbf{Fitting local optima}: The 2D reprojection objective can converge to local optima where joint configurations appear correct in the image plane but are inverted in 3D, as illustrated in \cref{fig:failed_case}. Incorporating complementary 3D constraints (\eg, triangulating correspondences across views) could help resolve such depth ambiguities. \textbf{Inference speed}: Without hand refinement or test-time scaling, our method takes $\sim$53s per subject (3s proxy generation + 50s fitting), comparable to EasyMoCap ($\sim$40s) while already achieving superior precision and robustness (\cref{tab:main_results}, ``w/o HR, TTS''). Hand refinement and test-time scaling provide further gains at additional cost. Future work could accelerate fitting by initializing with end-to-end predictions, and speed up proxy generation via consistency models~\cite{song2023consistency} or diffusion distillation. \textbf{Multi-view requirement}: Our method requires multiple views for reliable results, as single-view performance suffers from depth ambiguity. \textbf{Single-subject}: Extension to multi-person scenarios is straightforward by incorporating per-instance segmentation, with the primary challenge being cross-view identity association.

%% file: sec/6_conclusion.tex
\section{Conclusion}
\label{sec:conclusion}

We showed that the core of fitting-based human mesh recovery---predicting dense pixel-to-surface correspondences from images---is an image-to-image translation task well-suited for modern diffusion models. By training on large-scale synthetic data with pixel-perfect annotations and leveraging the visual priors of Stable Diffusion, DiffProxy overcomes the two bottlenecks that hindered prior dense-proxy approaches: noisy manual labels and limited generalization. Combined with hand refinement for fine-grained detail and test-time scaling for uncertainty-guided fitting, the resulting dense proxies provide uniform surface constraints that enable precise fitting. Trained exclusively on synthetic data without dataset-specific tuning, DiffProxy achieves state-of-the-art performance across five diverse real-world benchmarks, suggesting that the dense proxy route is a highly effective paradigm for precision-oriented tasks.

%% file: sec/7_acknowledgments.tex
\section{Acknowledgments}
\label{sec:acknowledgments}
This work was supported by the National Science Fund of China under Grant Nos. U24A20330,
62361166670, and 62376121, Basic Research Program of Jiangsu under Grant No. BK20251999,
Gusu Innovation Leading Talent Program under Grant No. ZXL2025319, and Jiangsu Provincial
Science \& Technology Major Project under Grant No. BG2024042.